\documentclass[conference]{IEEEtran}
\IEEEoverridecommandlockouts
\usepackage{cite}
\usepackage{amsmath,amssymb,amsfonts}
\usepackage{algorithmic}
\usepackage{graphicx}
\usepackage{textcomp}
\usepackage{xcolor}
\usepackage{placeins}

\title{
 Toward Study on the Effects of Robot's Humorous Apology on Human's Trust Recovery}

\author{Gayoung Kim$^1$, Go-Eum Cha$^2$ and Wonse Jo$^{1,*}$%
    \thanks{$^1$ A+HRI Laboratory, Department of Information and Telecommunication Engineering, Incheon National University, Incheon, South Korea. {\tt\small \{7rdud36, jow\}@inu.ac.kr}}
    \thanks{$^2$ Department of Computer Science, Purdue University, IN, USA. {\tt\small cha20@purdue.edu}}
    \thanks{$*$: corresponding author}%
    \thanks{This work was supported by the Ministry of Education of the Republic of Korea and the National Research Foundation of Korea (NRF-2026S1A5A8012989).}

}

\begin{document}
\maketitle

\begin{abstract}
Robot errors can disrupt human-robot collaboration by reducing user trust and willingness to rely on the robot. Robot apologies can support trust recovery after such errors, but their effects may vary with apology strategy and error severity. This study therefore presents a user experiment based on a 2$\times$2 between-subjects factorial design to examine the effect of humorous robot apologies on user trust recovery and the moderating role of error severity. The design crosses robot error severity (low \textit{vs.} high) with apology strategy (generic \textit{vs.} humorous). An online experiment will be conducted using the open-source Warehouse Robot Interaction Simulator. Participants will complete a collaborative task, experience a robot error followed by an apology, and report trust, trust recovery, interaction experience and reuse intention. We hypothesize that humorous apologies will produce greater trust recovery than generic apologies after low-severity errors, but that the advantage of humor will weaken or reverse after high-severity errors.
\end{abstract}
\vspace{0.5em}

\section{Introduction}
Human-robot collaboration (HRC) is rapidly expanding across logistics, manufacturing, and services, with robots functioning alongside humans as teammates. Sustained collaboration depends on users' trust in robots, yet robot errors can erode user trust and reduce users' willingness to rely on a robot in subsequent interactions.
According to previous studies \cite{esterwood2021you,zhang2023effects}, trust-repair strategies can be categorized into apology, denial, explanation, and promise, but the strategies have focused on formal, standardized apologies \cite{esterwood2021you,zhang2023effects}. Robot humor provides a different way to deliver apologies and may lead users to interpret the robot failure differently. The effect of apologies delivered through humor may also vary with robot error severity.
This study therefore explores the main effects of robot apology type and robot error severity on user trust recovery, as well as their interaction.

\section{Related Work}
Trust influences a human's belief that an automated system can help achieve their goals under uncertainty, and it is slow to build but can be sharply degraded by robot errors \cite{lee2004trust}. Prior HRI research has examined trust-repair strategies such as apology, denial, explanation, and promise \cite{esterwood2021you,zhang2023effects}; however, the role of humor in robot apologies remains insufficiently understood. Recent work on service robots suggests that humorous responses to robot failures can alleviate negative perceptions of failure and improve user attitudes \cite{yang2022exploring,zhang2025self,yu2025influence}. 
However, prior work has examined humorous \textit{versus} rational responses to service robot failures in one-time service encounters, with customer evaluation as the outcome \cite{yang2022exploring}. Such comparisons do not address how generic and humorous \textit{apologies} affect trust recovery during ongoing HRC.

In this paper, we propose an online, simulator-based user study that addresses this gap by examining the effect of a robot's humorous apologies on humans' trust recovery in a collaborative warehouse task.

\begin{figure}[!t]
    \centering
    \includegraphics[width=0.93\linewidth]{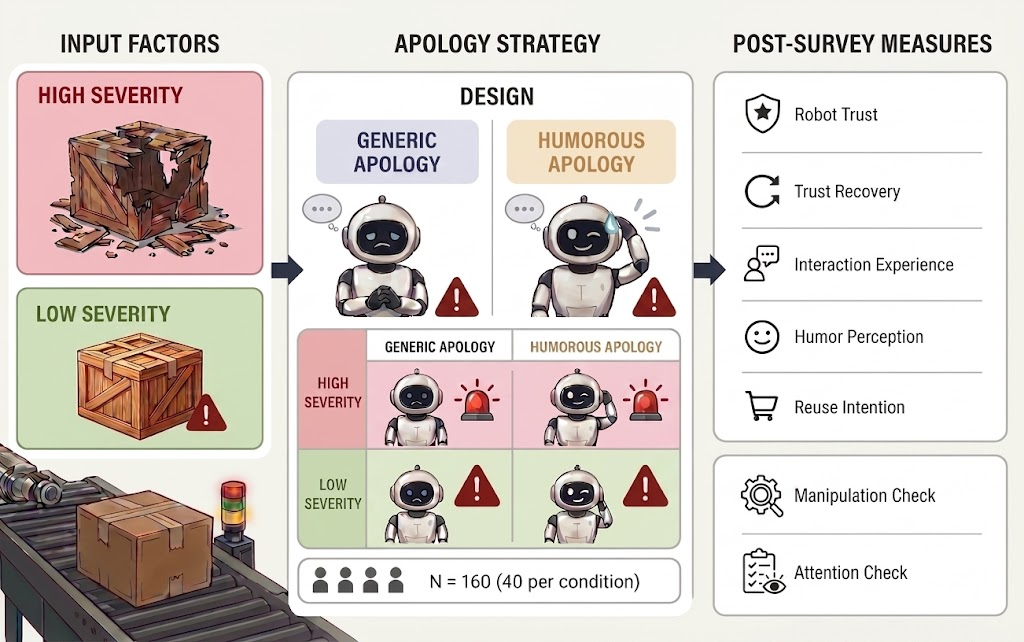}
    \vspace{-10pt}
    \caption{Conceptual overview of the hypothesized interaction between apology strategy and error severity on trust recovery. This figure was generated by ChatGPT and Google Gemini (image generation feature)}
    \label{fig:overview}
\end{figure}

\section{Design of Experiment}
The main objective of this study is to examine whether the trust-repair effect of a humor-based apology is not fixed but varies with error severity. While humor generally relieves tension and aids relationship repair, it may be perceived as trivializing the problem and impeding trust recovery when misaligned with context. Error severity is thus expected to moderate which of these effects predominates.

\begin{itemize}
\item $H_{1}$: The type of apology strategy will produce a significant difference in the level of user trust recovery.
\item $H_{2}$: The severity level of the error will produce a significant difference in the level of user trust recovery.
\item $H_{3}$: There will be a significant interaction effect between apology strategy and error severity. 
\end{itemize}

Specifically, a humorous apology is expected to be more effective under low, versus high, error severity.

\subsection{Experimental Design}
This study adopts a 2$\times$2 between-subjects factorial design with error severity (high \textit{vs.} low) and apology strategy (generic \textit{vs.} humorous) as independent variables, yielding four conditions (no control group included):
\begin{enumerate}
    \item Low error severity $\times$ generic apology
    \item Low error severity $\times$ humorous apology
    \item High error severity $\times$ generic apology
    \item High error severity $\times$ humorous apology
\end{enumerate}

\indent The experiment uses the open-source Warehouse Robot Interaction Simulator \cite{esterwood2023warehouse}, in which a robot presents boxes with visible identification numbers and participants respond by clicking on-screen O/X buttons to confirm or reject each box. Error severity will be operationalized as the magnitude of disruption caused by the robot's error: a delayed box delivery accompanied by a minor score deduction (low severity) versus retrieval of an incorrect box accompanied by a larger score deduction (high severity); specific parameters are being finalized. Humorous apology candidates, combining humor type, facial expression, gesture, and voice, are validated via a preliminary survey; the highest-rated candidate is used in the main experiment.

\subsection{Experimental Procedure}
Participants complete five stages online: (1) consent; (2) pre-survey (demographics including age, gender, nationality, and prior experience with robots/AI; personality traits); (3) a trial task familiarizing participants with the simulator interface; (4) the main experiment, in which participants are automatically and randomly assigned to one of four experimental conditions during a collaborative box-sorting task, and an error occurs per condition followed by a generic or humorous apology; and (5) outcome assessment (such as, trust, trust recovery, interaction experience, reuse intention, perceived error severity), including an attention check.

\subsection{Measures}
All measures in the present study were assessed using a 7-point Likert scale and were constructed with reference to prior research. The robot trust scale was adapted from \cite{lee2004trust}; trust recovery items were constructed with reference to prior trust-repair literature; interaction experience items were constructed with reference to prior HRI research; humor perception was assessed based on participants' perceived humor of the robot's apology. Measured variables include (1) robot trust, (2) trust recovery, (3) interaction experience, (4) humor perception, (5) reuse intention, and (6) manipulation check for error severity.

\subsection{Analysis Plan}
An a priori power analysis (G*Power 3.1.9.7; F tests, ANOVA: fixed effects, main effects and interactions; f = 0.25, $\alpha$ = .05, power = .80, numerator df = 1, groups = 4) indicated a minimum required sample size of 128. Accounting for an expected 20\% attrition rate, we plan to recruit 160 human subjects (40 per condition) via online research panel services.
Data will be analyzed using a 2$\times$2 two-way ANOVA, testing the main effect of apology strategy ($H_{1}$), error severity ($H_{2}$), and their interaction ($H_{3}$).

\subsection{Expected Contribution}
By empirically identifying the boundary conditions under which humor-based apology aids or hinders trust recovery, this study is expected to inform the design of context-sensitive robot communication strategies for HRC environments.

\begin{figure}[!t]
    \centering
    \vspace{3pt}

    \includegraphics[width=0.95\linewidth,height=5cm,keepaspectratio]{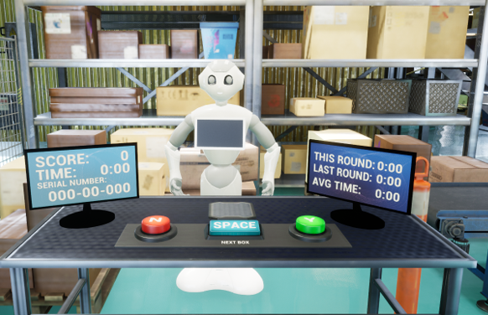}
    \vspace{-5pt}
    \caption{Main screen of the warehouse simulator \cite{esterwood2023warehouse}.}
    \label{fig:placeholder}
\end{figure}

\section{Conclusion and Future Works}
This paper presents a user experiment design using a 2$\times$2 factorial framework to explore when humor-based apologies either aid or hinder trust recovery in HRC scenarios by manipulating error severity and apology strategy.

In future work, we plan to explore whether the effectiveness of humorous apologies varies with error severity through online experiments. Additionally, we aim to extend this framework to include a more diverse participant population and examine the mediating role of humor perception in the relationship between apology strategy and trust recovery.

\bibliographystyle{IEEEtran}
\bibliography{references}

\end{document}